\documentclass{article}

\PassOptionsToPackage{numbers, compress}{natbib}

\usepackage[preprint]{neurips_2026}

\usepackage[utf8]{inputenc} 
\usepackage[T1]{fontenc}    
\usepackage{url}            
\usepackage{booktabs}       
\usepackage{amsfonts}       
\usepackage{nicefrac}       
\usepackage{microtype}      
\usepackage{xcolor}         
\usepackage{graphicx}
\usepackage{booktabs}
\usepackage{bbm}
\usepackage{algorithm}
\usepackage{algorithmic}
\definecolor{RefBlue}{RGB}{46, 125, 190}
\usepackage[colorlinks=true,
            citecolor=RefBlue,
            linkcolor=RefBlue,
            urlcolor=RefBlue]{hyperref}
\usepackage{multirow} 
\usepackage{soul}
\usepackage{url}
\usepackage{anyfontsize}
\usepackage{booktabs}
\usepackage{algorithm}
\usepackage{algorithmic}
\usepackage{amsfonts}
\usepackage{amssymb}
\usepackage{indentfirst}
\usepackage{bbding}
\usepackage{colortbl}
\usepackage{amsmath}
\usepackage{caption}
\usepackage{tikz}
\usepackage{wrapfig} 
\usepackage{enumitem} 
\usepackage{changepage}  
\definecolor{customgreen}{rgb}{0.0, 0.7, 0.3}
\usepackage{amsmath}      
\usepackage{xcolor}       
\usepackage{tcolorbox}    
\usepackage{changepage}   

\title{OmniView-Space: Reinforcing Spatial Reasoning via Multi-Perspective Spatial Mapping}

\author{%
  Xudong Li$^{1*}$ \quad 
  Mengdan Zhang$^{2}$\thanks{Equal Contribution.} \quad 
  Peixian Chen$^{2}$ \quad 
  Jiaxi Tan$^{1}$ \quad 
  Zihao Huang$^{3}$ \quad 
  \\ 
  \textbf{Jingyuan Zheng}$^{1}$ \quad 
  \textbf{Yan Zhang}$^{1}$ \quad 
  \textbf{Xiawu Zheng}$^{1}$ \quad 
  \textbf{Xing Sun}$^{2}$ \quad 
  \textbf{Rongrong Ji}$^{1}$
  \\
  \\
   $^{1}$ Key Laboratory of Multimedia Trusted Perception and Efficient Computing, \\
    Ministry of Education of China, Xiamen University, 361005, P.R. China\\
    $^{2}$ Tencent Youtu Lab \quad
    $^{3}$ Beijing Institute of Technology\\
  \texttt \small{\{lxd761050753, zhangmengdanrz\}@gmail.com, \{zhengxiawu, rrji\}@xmu.edu.cn}\\
}

\begin{document}

\maketitle

\vspace{-20pt}
\begin{abstract}
Spatial intelligence remains a persistent challenge for Multimodal Large Language Models (MLLMs), as it requires coherent spatial scene representations beyond basic object recognition. Existing methods typically build such representations through textual reasoning or 3D reconstruction. However, they often falter during multi-step reasoning, particularly when required to dynamically re-anchor evidence to the specific camera-, object-, or direction-centric reference frames demanded by complex queries. To address this, we propose \textbf{OmniView-Space}, a framework designed to maintain spatial consistency through multimodal egocentric evidence. Our approach consists of three core components: (1) {Multi-Perspective Spatial Mapping (MPSM)}, which re-anchors reconstructed geometry into a query-aligned visual cognitive map and a textual spatial graph; (2) {Tool-Guided Egocentric Reasoning}, an interleaved policy trained to actively select the ego anchor required by the query and request the corresponding MPSM evidence; and (3) {Cognitive-Map Distillation}, which uses MPSM-generated trajectories and ego-frame rewards to train the model to reason with self-generated cognitive maps. Experiments on single- and multi-image spatial reasoning benchmarks show that OmniView-Space achieves state-of-the-art performance. Furthermore, the distilled model maintains this performance while reducing reliance on external geometry pipelines.
\end{abstract}
\newcommand{\ovsSmallCircle}[1]{\tikz[baseline=(c.base)]{\node[draw,circle,inner sep=0.5pt,line width=0.75pt,font=\scriptsize] (c) {#1};}}
\newcommand{\ovsBlackCircle}[1]{\tikz[baseline=(c.base)]{\node[fill=black,circle,inner sep=0.65pt,font=\scriptsize,text=white] (c) {#1};}}
\section{Introduction}

Spatial intelligence refers to the ability to reason about the geometric and physical structure of the world from visual observations. Although Multi-modal Large Language Models (MLLMs) have made substantial progress in general visual understanding~\citep{llava,llava-ov,qwen2vl,chen2024internvl,qwen3-vl}, they still struggle with spatial questions involving relative localization, viewpoint taking, cross-view association, or camera-motion reasoning~\citep{vsi,mmsi,spar}. Unlike conventional visual question answering, which mainly focuses on semantic recognition in static images, spatial intelligence requires maintaining a consistent spatial state throughout the reasoning process. The challenge is therefore not only to recognize scene content, but also to preserve spatial consistency as intermediate reasoning steps unfold. For example, when asked what lies to the right of a chair from the point of view of the chair, knowing the room layout is insufficient. The model also needs to establish a chair-centered spatial state and reason within that reference frame; otherwise, judgments regarding left and right can easily become confused.
\begin{figure}[t]
\centering
\includegraphics[width=0.98\linewidth]{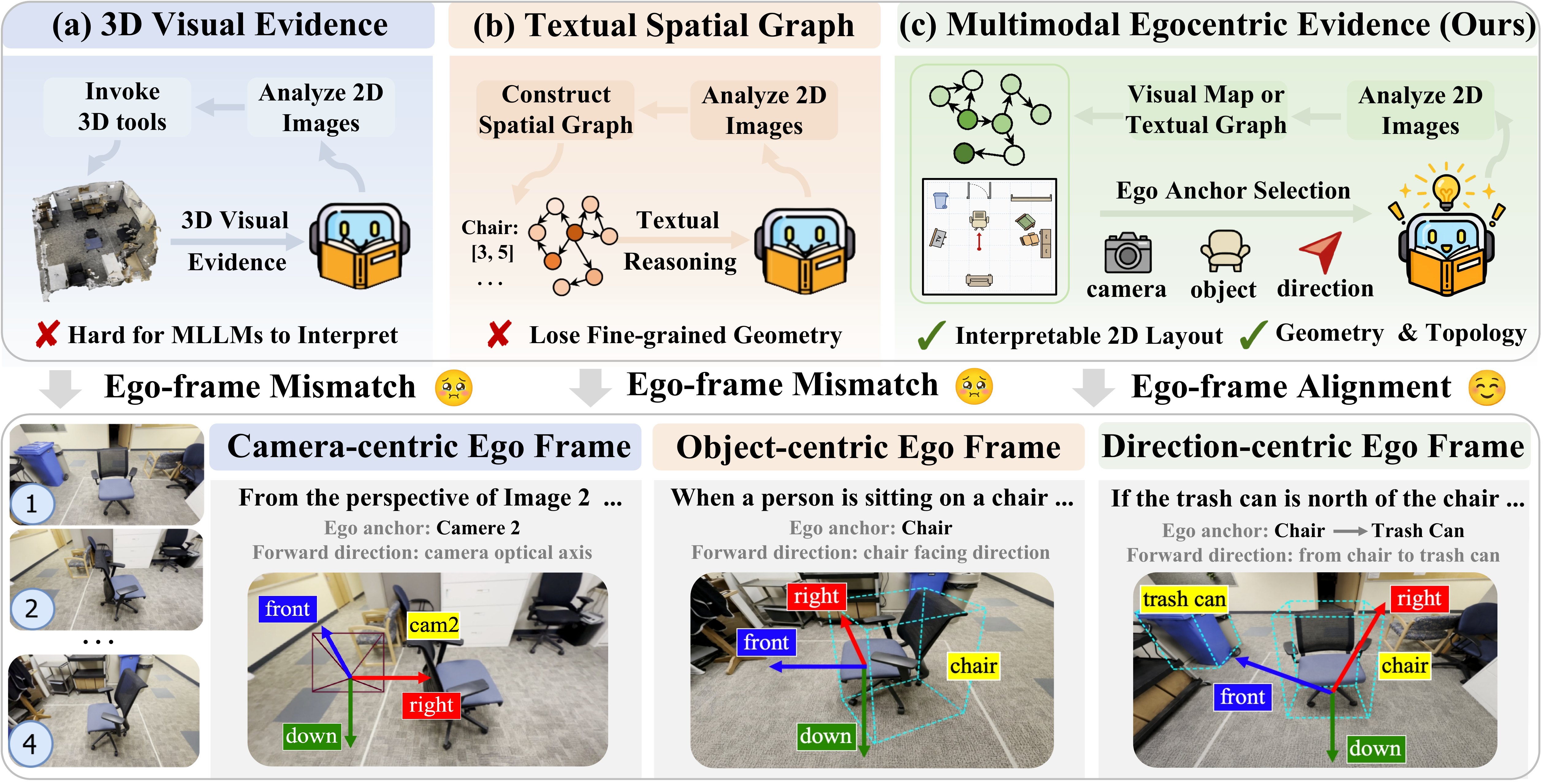}
\caption{(a) 3D visual evidence preserves the metric structure, but raw reconstruction outputs are difficult for MLLMs to interpret directly. (b) Textual spatial graphs are language-accessible, but coordinate-style reasoning can lead to the loss of fine-grained geometry. (c) OmniView-Space builds multimodal egocentric evidence: a visual BEV cognitive map and a textual spatial graph aligned to the query-specified ego frame. Bottom: existing evidence can be frame-mismatched, while OmniView-Space aligns it to the camera-, object-, or direction-centric frame required by the question.}
\label{fig:overview}
\vspace{-20pt}
\end{figure}

While textual chain-of-thought excels in symbolic and mathematical domains~\citep{grpo,ouyang2025spacer}, it is less reliable for spatial intelligence~\citep{cai2025scaling,vst}. Purely textual traces often distort object correspondence, relative orientation, and topology, causing spatial errors to compound. To provide MLLMs with stronger spatial grounding, recent work explores two main forms of intermediate evidence (Fig.~\ref{fig:overview}(a,b)). One form injects 3D visual evidence through 3D-aware inputs or external reconstruction~\citep{luo2026pyspatial,think3d}. It preserves metric structure, but raw 3D outputs are often challenging for MLLMs to interpret. The other form constructs textual spatial graphs or grid maps~\citep{vsi,huang2025video2layout,mindcube}. It makes relations more language-accessible, but coordinate-style reasoning can lose fine-grained geometry and topology. 
More importantly, both forms usually express evidence in a system-chosen reference frame (e.g., the first camera view or a world system), creating a mismatch with many spatial questions that require a \emph{query-specified ego frame}. As shown in Fig.~\ref{fig:overview}, individual questions anchor this frame to distinct \emph{ego anchors}, such as cameras, objects, or directions. However, existing methods rely on textual coordinate representations to translate system-frame evidence into the query-specified frame, where performing such a geometric transformation purely through text is highly unreliable and easily breaks spatial consistency during multi-step reasoning.

Humans handle such ego-frame shifts naturally: we can keep a coherent mental scene, imagine it from the query-specified ego frame, and then make a spatial judgment. In contrast, current MLLMs often struggle when the query-specified ego frame differs from the observed camera or system-defined frame~\citep{3dsrbench,zhang2024vision}.
To address this challenge, we propose \textbf{OmniView-Space}, a framework for multimodal egocentric evidence construction and reasoning, as shown in Fig.~\ref{fig:overview}(c). This design reframes diverse camera-, object-, and direction-centric queries as reasoning in the query-specified ego frame, following the intuition that if a model can perceive the scene in this frame, spatial reasoning becomes easier to ground. The framework uses two complementary forms of MPSM evidence: a \emph{visual BEV cognitive map} for inspectable 2D layout and a \emph{textual spatial graph} for explicit relational structure. It follows a progressive design that moves from external spatial evidence to tool-using reasoning and finally to self-generated cognitive-map reasoning. \ovsSmallCircle{1} First, we develop \emph{Multi-Perspective Spatial Mapping} (MPSM), which constructs multimodal egocentric evidence from multi-view images, task-relevant objects, and the ego anchor. MPSM returns both forms of evidence in the same query-specified ego frame, making object layout, relative direction, and topology directly usable for reasoning. \ovsSmallCircle{2} Second, we train an interleaved tool-using policy with reinforcement learning (RL). The model learns to analyze the question, identify the ego anchor, request either the visual map or textual graph, and ground its answer in the returned multimodal egocentric evidence. \ovsSmallCircle{3} Finally, we distill this tool-guided behavior into the model itself. MPSM-generated trajectories teach the model to predict an egocentric cognitive map before answering, and during GRPO~\citep{grpo}, MPSM further serves as a reward reference for refining the predicted objects, camera layout, and final answer.

Experiments on multi-image and single-image spatial reasoning benchmarks show that OmniView-Space significantly improves spatial reasoning over strong MLLM baselines. Notably, the distilled model closes much of the gap to the tool-augmented pipeline while removing the need to invoke the full geometry pipeline at inference time. Our contributions are summarized as follows:
\vspace{-5pt}
\begin{itemize}
\item[\ovsBlackCircle{1}] \emph{Query-aligned egocentric evidence.} We identify reference-frame mismatch as a bottleneck in multi-step spatial reasoning: evidence is often expressed in a system-defined frame, while spatial questions require judgment from a camera-, object-, or direction-centric ego frame. This motivates query-aligned egocentric evidence as the interface for spatial reasoning.

\item[\ovsBlackCircle{2}] \emph{Multi-perspective spatial mapping.} We introduce the MPSM toolkit, which converts scene observations into multimodal egocentric evidence: a query-aligned visual BEV cognitive map and a textual spatial graph that combine 2D interpretability with explicit geometry.

\item[\ovsBlackCircle{3}] \emph{Tool-guided reasoning and distillation.} We train an interleaved tool-using policy with RL to select the ego anchor required by the query and actively call MPSM for visual cognitive map or textual spatial graph construction. We then distill this tool-guided reasoning into self-generated cognitive maps using MPSM-generated trajectories and ego-frame rewards.
\end{itemize}

\section{Related Work}

\noindent\textbf{Spatial reasoning in MLLMs.}
Spatial reasoning remains a foundational challenge for Multi-modal Large Language Models (MLLMs)~\citep{jia2025omnispatial,chow2025physbench,vst}. Recent progress has been driven by stronger MLLMs~\citep{qwen3-vl,qwen2vl,llava-ov}, broader benchmarks~\citep{spar,mmsi,vsi,3dsrbench}, and training or prompting strategies that use 3D reconstruction, depth cues~\citep{roy2025bydeway}, large-scale 3D spatial VQA data~\citep{balazadeh2024synthetic,zhang2025spatial}, spatial prompts~\citep{30,liu2025coarse,marsili2025visual,taguchi2025spatialprompting}, mental simulation~\citep{chen2025think,leroy2024grounding}, visual chain-of-thought or RL-based reasoning~\citep{fan2025grit,shao2024visual,wang2025visuothink,30}, and explicit visual grounding~\citep{ViLaSR}. In multi-view settings, multiple observations can help recover 3D structure and reduce single-view ambiguity, and multiview-augmented MLLMs~\citep{11,4,30,wu2025spatial,68,85,86} improve geometric understanding and perspective taking. However, these advances still leave a gap in maintaining spatial evidence in the query-specified ego frame, especially when the answer must be judged from a camera-, object-, or direction-centric perspective rather than a system-defined frame.

\noindent\textbf{Geometry-augmented tool use.}
Tool calling strengthens MLLMs by invoking external specialists through prompting, code generation, or learned policies~\citep{chen2025geometrically,think3d,luo2026pyspatial,ViLaSR,han2025tiger}. In spatial domains, these tools provide depth, segmentation, 3D reconstruction, camera poses, or viewpoint exploration as intermediate evidence~\citep{da3,sam3,vggt,think3d}. Other approaches inject 3D-aware features or camera-guided representations directly into the network~\citep{chen2025think,spacemind}. While demonstrating the value of explicit geometry, raw 3D outputs are often hard for MLLMs to interpret, widening the gap between geometry and language-level reasoning. Furthermore, their evidence remains fixed to a system-defined frame, motivating our approach to treat geometry as callable evidence aligned with the query's ego anchor.

\noindent\textbf{Cognitive maps and frame grounding.}
Spatial cognition includes mental rotation, spatial visualization, and object assembly~\citep{64,69,77}, and is often linked to Spatial Mental Models (SMMs)~\citep{26,27}. Recent MLLM studies probe these abilities via coordinate-aware prompting, CoT reasoning, representation alignment, and RL~\citep{6,39,42,7,10,45}. A related cognitive-map paradigm uses coordinate or grid representations for spatial reasoning~\citep{vsi,egomind,spatialthinker,ouyang2025spacer,huang2025video2layout}. However, many maps are discrete, fixed to a world or first-camera frame, or limited to single-frame inputs~\citep{4,42}, restricting their utility when answers depend on a query-specified ego frame. Recent allocentric-reasoning work similarly notes that MLLMs struggle when a query requires instantiating a target frame rather than using the observed view~\citep{wang2026allocentric}. OmniView-Space addresses this by constructing multimodal egocentric evidence—including a visual BEV cognitive map and a textual spatial graph—and internalizing tool-guided reasoning through MPSM trajectories and ego-frame rewards.

\section{Methodology}
\begin{figure*}[t!]
\centering
\includegraphics[width=138mm,height=52mm]{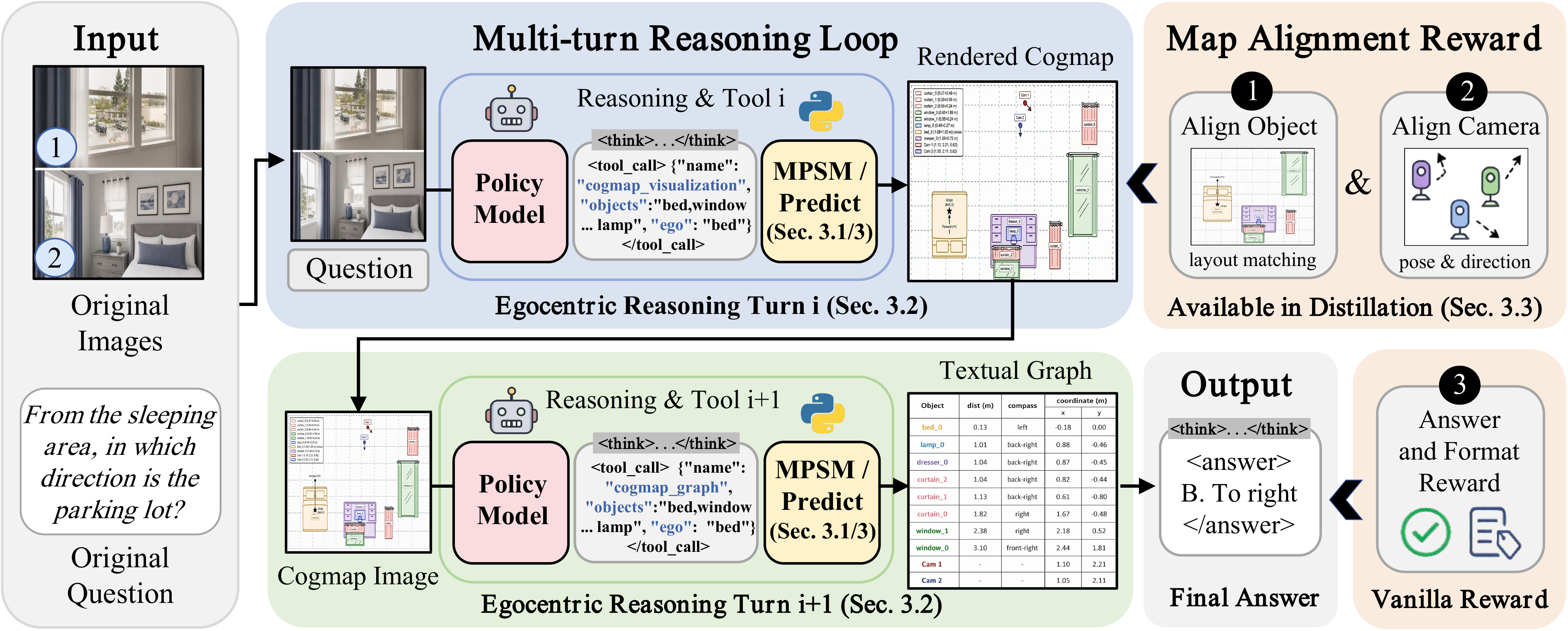}
\caption{\textbf{Overview of the OmniView-Space.} Given a question and an image, the policy analyzes the ego anchor to either invoke MPSM (Sec.~\ref{sec:toolkit}) or directly predict (Sec.~\ref{sec:distill}) a query-aligned visual BEV cognitive map or textual spatial graph across multiple reasoning turns, where this trajectory is optimized via reinforcement learning (RL) solely using an answer reward. During the distillation phase, MPSM generates trajectories for cold-start training and provides a reward for map alignment during RL, which instructs the model to reason using internally generated cognitive maps (Sec.~\ref{sec:distill}).}
\label{fig:method_overview}
\vspace{-15pt}
\end{figure*}

\definecolor{OVSToolYellow}{HTML}{FFF2CC}
\newcommand{\ovsTool}[1]{\begingroup\setlength{\fboxsep}{1.2pt}\colorbox{OVSToolYellow}{\texttt{#1}}\endgroup}
Given a question $q$ and images $\mathcal{I}$, the proposed OmniView-Space grounds spatial reasoning in evidence expressed in the ego frame required by the query. We first define MPSM as an external evidence constructor (Sec.~\ref{sec:toolkit}). As shown in Fig.~\ref{fig:method_overview}, the model then analyzes the question and images, selects the ego anchor required by the query, calls MPSM for a query-aligned visual BEV cognitive map or textual spatial graph, and continues reasoning over the returned evidence (Sec.~\ref{sec:rl}). Finally, we distill this tool-guided process into the model itself, so that it can reason with self-generated query-aligned cognitive maps without calling the full geometry pipeline at inference (Sec.~\ref{sec:distill}).

\subsection{Multi-Perspective Spatial Mapping}\label{sec:toolkit}
\noindent\textbf{Role and interface.}
As shown in Fig.~\ref{fig:mpsm_toolkit}, MPSM is an evidence constructor that takes the multi-view images $\mathcal{I}$, task-relevant object hints $\mathcal{H}$, and an ego anchor $r$ selected by the MLLM as input. The hints $\mathcal{H}$ can be object names and 2D boxes, while $r$ can be an input camera, an object with an estimated facing direction, or a direction defined by a source-to-target object relation. The MPSM operator $\Omega$ returns two forms of the same query-aligned scene evidence: \ovsTool{cogmap\_visualization} renders a visual BEV cognitive map $\Pi_r$, while \ovsTool{cogmap\_graph} exports a textual spatial graph $\mathcal{T}_r$ with object attributes, containment/support edges, distances, and ego-relative directions. Formally,
\begin{equation}
\label{eq:toolkit}
\mathcal{G}_r=\Omega(\mathcal{I},\mathcal{H},r)
:=\bigl\{r,\Pi_r,\mathcal{T}_r\},
\end{equation}
where $\mathcal{G}_r$ is the spatial state re-anchored to ego reference $r$, $\Pi_r$ and $\mathcal{T}_r$ are the rendered BEV cogmap and the relation graph expressed in the same ego frame. Here, the index $r$ denotes outputs transformed into the ego frame (Eq.~\ref{eq:egoframe}). To construct this spatial state, MPSM performs two main steps:

\begin{wrapfigure}{r}{0.48\linewidth}
\vspace{-10pt}
\centering
\includegraphics[width=\linewidth]{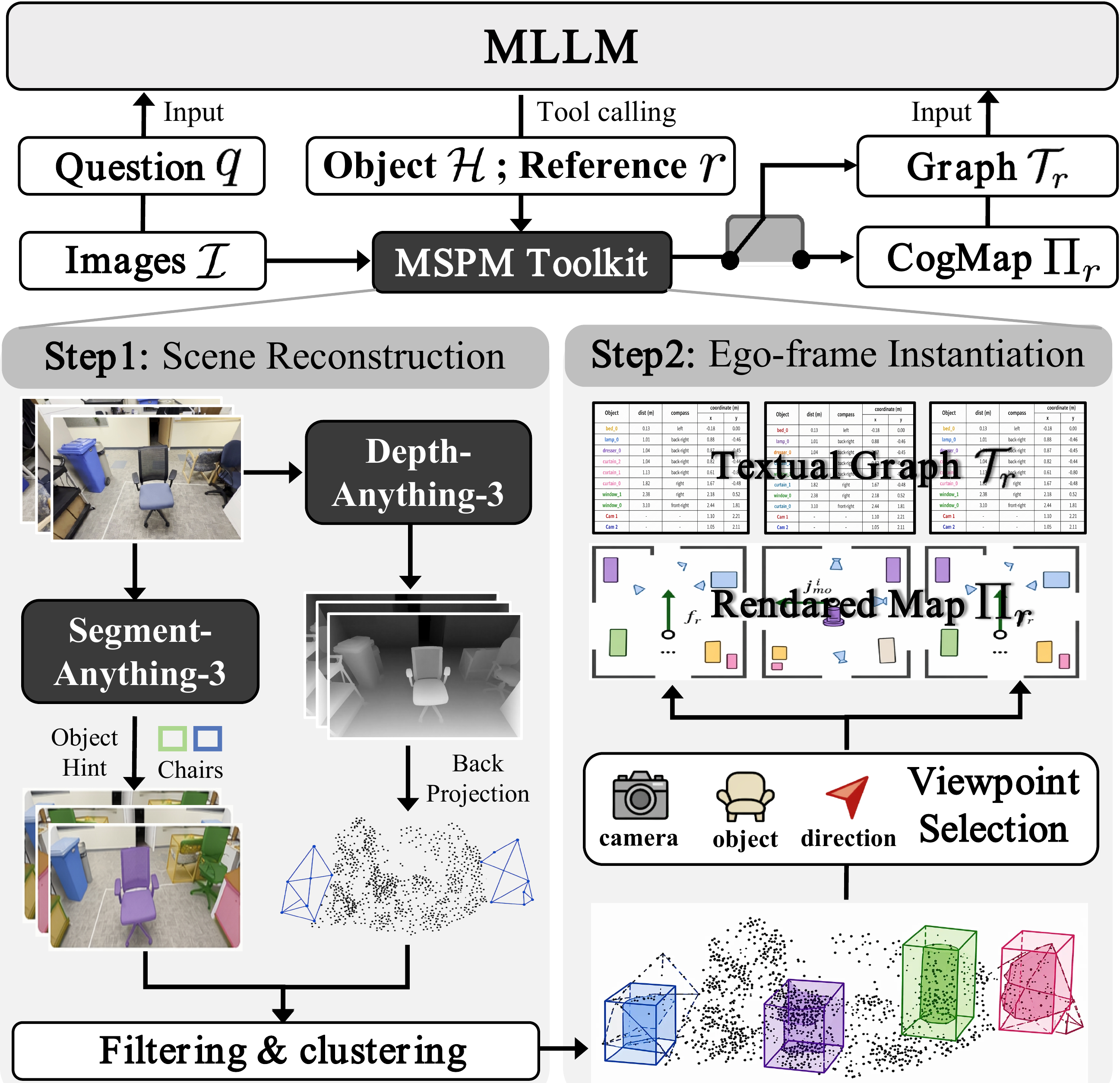}
\caption{\textbf{MPSM toolkit.} Given multi-view images, object hints, and an ego reference, MPSM reconstructs a metric scene, re-anchors it to the query-specified frame, and returns both a rendered BEV cogmap and a textual spatial graph.}
\label{fig:mpsm_toolkit}
\vspace{-10pt}
\end{wrapfigure}
\noindent\textbf{Step 1: Scene Reconstruction in the default reference frame.}
To obtain metric evidence, MPSM reconstructs a 3D scene from the input views. For each image $I_t$ of resolution $H\times W$ and pixel $(u,v)$, a dense geometry model~\citep{da3} predicts a point map $X_t$, a validity mask $U_t$, a confidence map $C_t$, and the camera pose. A promptable segmenter~\citep{sam3} grounds the object hints $\mathcal{H}$ into foreground masks, using text prompts, boxes, or both depending on the available hint type. We keep only pixels and frames that pass a confidence gate,
\begingroup
\setlength{\abovedisplayskip}{4pt}
\setlength{\belowdisplayskip}{2pt}
\begin{equation}
\label{eq:conf_gate}
\begin{aligned}
V_t &= \mathbf{1}[U_t{=}1]\cdot\mathbf{1}[C_t{>} \tau_{\mathrm{pix}}]\cdot\mathbf{1}[\bar C_t{>}\tau_{\mathrm{frm}}],\\
\bar C_t &= \frac{1}{HW}\sum_{u,v}C_t(u,v).
\end{aligned}
\end{equation}
\endgroup
where $V_t$ is the retained-pixel mask, $\bar C_t$ is the average frame confidence, and $\tau_{\mathrm{pix}}$ and $\tau_{\mathrm{frm}}$ are pixel- and frame-level thresholds. The retained foreground points are fused across views, voxelized, assigned semantic labels by majority vote, and clustered into object instances. Each instance is represented by a 3D box and BEV footprint, with its intrinsic facing direction estimated using~\citep{wang2026orient} when required for object-centric queries.

\noindent\textbf{Step 2: Ego-frame Instantiation.}
Since the reconstructed scene is initially expressed in the default reference frame, we should re-express it in the query-specified ego frame. After gravity alignment, the ego anchor $r$ defines an origin $o_r$ and a forward direction $f_r$ on the ground plane. The pair $(o_r,f_r)$ is instantiated differently for each anchor type: (1) a camera anchor uses the camera center and optical axis, (2) an object anchor uses the object centroid and estimated facing direction, and (3) a direction anchor uses a source-object centroid with the source-to-target vector as forward. Let $\tilde p\in\mathbb{R}^3$ be a point in the gravity-aligned scene coordinate system, and let $R_r$ be the yaw rotation that maps $f_r$ to the positive BEV $Y$ axis. The coordinate of $\tilde p$ in the query-specified ego frame is:
\begin{equation}
\label{eq:egoframe}
p^{(r)}=R_r(\tilde p-o_r).
\end{equation}
The same transformation is applied to object boxes, BEV footprints, and camera poses. MPSM renders $\Pi_r$ with the ego at the origin and exports $\mathcal{T}_r$ with the same ego-relative geometry, so the visual BEV cognitive map and textual spatial graph remain consistent views of one spatial state.

\subsection{Tool-Guided Egocentric Reasoning}\label{sec:rl}
To leverage MPSM's egocentric evidence, we train the MLLM as an interleaved tool-use policy. The model learns to analyze the query, select the required ego anchor and object hints, invoke the visual BEV cognitive map or textual spatial graph, and ground its answer in the returned spatial state.

\noindent\textbf{Interleaved rollout.}
At the decoding step $t$, the policy conditions on both generated text and multimodal observations returned by tools. We denote the rollout state and action distribution as
\begin{equation}
\label{eq:rollout}
s_t=\{X_{\le t};\,M_{\le t}\},\qquad a_t\sim\pi_\theta(a\mid s_t),
\end{equation}
where $X_{\le t}$ is the model-generated text history, $M_{\le t}$ contains the original images and any MPSM observations, $a_t$ is the next text or tool action, and $\pi_\theta$ is the MLLM policy with parameters $\theta$. Tokens from tool observations are masked from the policy gradient, so optimization is applied only to the model's analysis, tool parameters, reasoning, and answer.

\begin{wrapfigure}{r}{0.45\linewidth}
\vspace{-10pt}
\centering
\includegraphics[width=\linewidth]{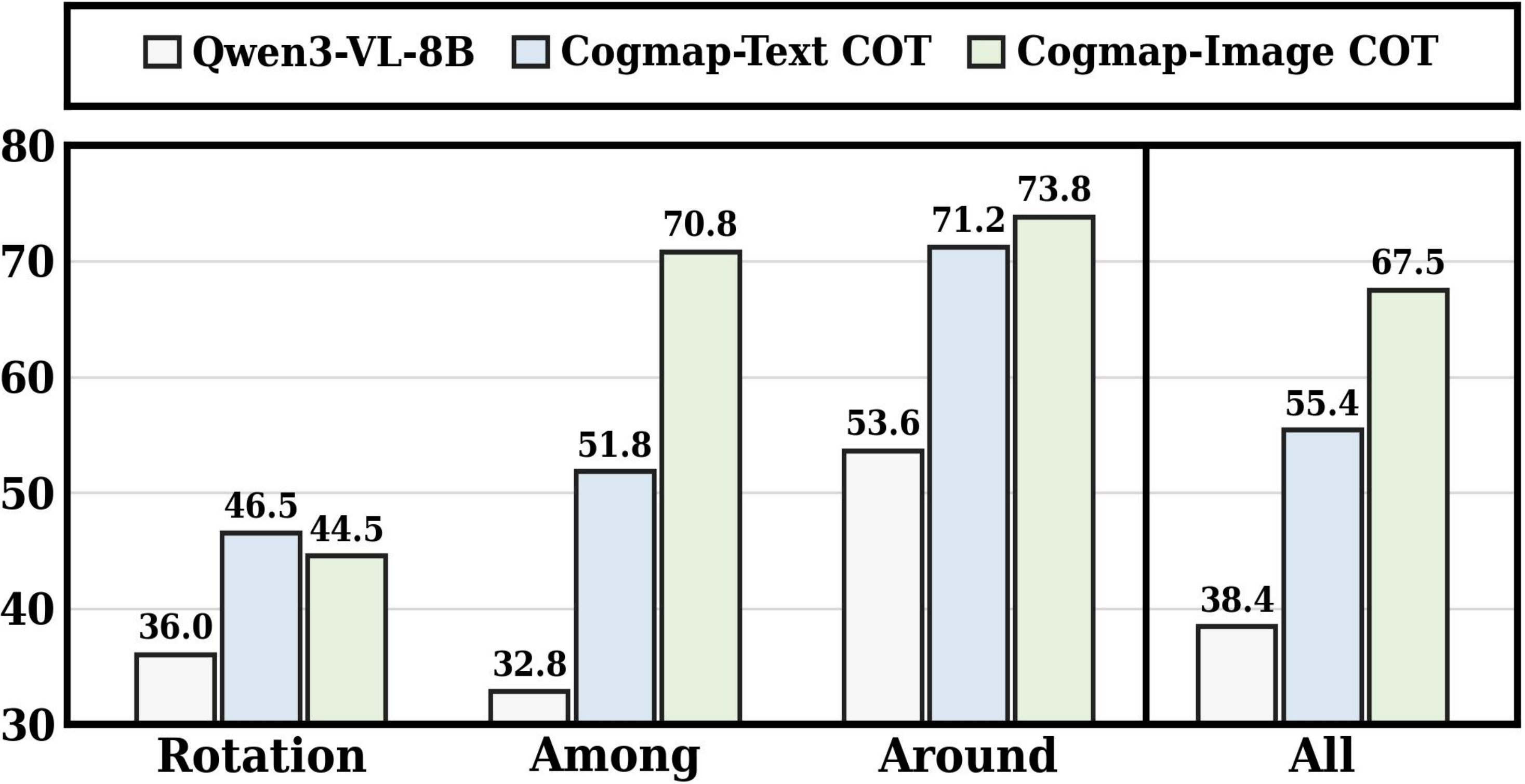}
\caption{\textbf{Comparison of evidence formats.} On MindCube, using a visual BEV cognitive map outperforms text-coordinate reasoning overall, suggesting that visualized egocentric evidence is a more usable intermediate representation for current MLLMs.}
\label{fig:visual_cogmap_interface}
\vspace{-10pt}
\end{wrapfigure}

\noindent\textbf{Complementary visual and textual evidence.}
Prior spatial reasoning methods often keep intermediate maps in textual or coordinate-based form, requiring the model to infer spatial relations from serialized layouts or JSON-style representations~\citep{vsi,mindcube}. We train and evaluate two variants on the MindCube benchmark, one using text-coordinate maps and the other using visual BEV cognitive maps. As shown in Fig.~\ref{fig:visual_cogmap_interface}, the visual-map condition achieves better overall performance, suggesting that visual layouts are more suitable for qualitative viewpoint reasoning such as left--right, front--back, and between-object judgments. The textual spatial graph remains complementary: it provides more explicit numerical and structural evidence for distance comparisons, height or support relations, and object attributes, which is consistent with the tool-call analysis in Fig.~\ref{fig:tool_distribution}, where graph calls become more frequent on 3D relation tasks. The original views remain reliable for visual identity. We therefore allow the policy to choose among the original images, the visual BEV cognitive map $\Pi_r$, and the textual spatial graph $\mathcal{T}_r$ according to the spatial demand of each query, rather than imposing a fixed modality schedule.

\noindent\textbf{Policy optimization.}
During a rollout, the model interactively invokes the MPSM tool---specifying the target hints, ego reference, and desired modality (visual map $\Pi_r$ or relation graph $\mathcal{T}$)---to gather spatial observations. For a complete trajectory $\tau$ terminating with a final answer $\hat{a}$, we optimize the MLLM policy via GRPO~\citep{grpo} with the reward objective:
\begin{equation}
\label{eq:tool_reward}
R_{\mathrm{tool}}(\tau)=R_{\mathrm{ans}}(\hat a)+\lambda_{\mathrm{fmt}}R_{\mathrm{fmt}}(\tau),
\end{equation}
where $R_{\mathrm{ans}}$ scores answer correctness, $R_{\mathrm{fmt}}$ enforces the tool-call and answer format~\citep{deepeyes}, and $\lambda_{\mathrm{fmt}}$=0.2 controls the strength of the format reward. This trajectory-level objective trains ego-frame selection, tool invocation, map/graph reading, and final answering jointly.

\subsection{Distilling Cognitive-Map Reasoning}\label{sec:distill}
The tool-using policy provides reliable egocentric evidence, but it still depends on external geometry tools. Next, we will explore how to distill the tool-guided reasoning ability into the model: given $(q,\mathcal{I})$, the model learns to predict a JSON-format ego-frame cognitive map, which is rendered into a visual BEV cognitive map for subsequent reasoning, as shown in Fig.~\ref{fig:method_overview}.

\noindent\textbf{MPSM-guided SFT Data.}
To enable the model to actively generate such cognitive maps without relying on external tools, we first collect expert trajectories that combine geometric tool outputs with reasoning traces. Specifically, we construct Supervised Fine-Tuning (SFT) trajectories, leveraging MPSM for intermediate cognitive mapping and Qwen3-VL-235B~\citep{qwen3-vl} for generating reasoning traces. For each sample, the MLLM first extracts relevant object hints and predicts the required ego anchor. Next, MPSM constructs a query-aligned visual BEV cognitive map (Sec.~\ref{sec:toolkit}), denoted as $J_s$. The MLLM then generates two rationales: $R_1$, which explains the construction of $J_s$, and $R_2$, which infers the predicted answer $\hat{a}_s$ from the inputs and $J_s$. The resulting SFT trajectory is assembled as: \texttt{<think>} $R_1$ \texttt{</think>} $\rightarrow$ \texttt{<cogmap>} $J_s$ \texttt{</cogmap>} $\rightarrow$ [\textit{Rendered Image}] $\rightarrow$ \texttt{<think>} $R_2$ \texttt{</think>} $\rightarrow$ \texttt{<answer>} $\hat a_s$ \texttt{</answer>}. Before training, we filter noisy trajectories via answer checking, critic-based repair, and missing-object refinement. Refer to the appendix for more details.

\noindent\textbf{Map Alignment Reward.}
SFT initializes the model to generate cogmaps and map-grounded rationales, but 1D token-level imitation does not explicitly reward 2D map geometric alignment. We therefore further train with GRPO using both answer rewards and map-quality rewards. Let $\hat J$ be the predicted JSON-format cogmap and $\hat a$ be the final answer. The total reward is defined as:
\begin{equation}
\label{eq:internal_reward}
R(\hat J,\hat a)=R_{\mathrm{ans}}(\hat a)+\lambda_{\mathrm{fmt}}R_{\mathrm{fmt}}(\hat J,\hat a)+g(\hat a)\Bigl(
\alpha \cdot R_{\mathrm{object}}(\hat J)
+ \beta \cdot R_{\mathrm{cam}}(\hat J)
\Bigr),
\end{equation}
where $R_{\mathrm{ans}}$ and $R_{\mathrm{fmt}}$ evaluate answer correctness and format compliance, $\alpha$, $\beta$ are hyper-parameters balancing different reward components, and $g(\hat a)=\mathbf{1}[\hat a=a^{\ast}]$ applies map alignment only to correct predictions. The alignment term has two parts. (1) \textbf{Object alignment} $R_{\mathrm{object}}=\sum_{c\in\mathcal{C}}\frac{n_c}{N_o}R_c$, which first performs category-wise Hungarian matching in the query-specified ego BEV frame, then computes $R_c$ from matched object pairs using BEV bbox center accuracy and size consistency. (2) \textbf{Camera alignment} $R_{\mathrm{cam}}=\frac{1}{N_c}\sum_i(r_{\mathrm{pos}}^i+r_{\mathrm{ori}}^i)$ measures camera position, pairwise direction consistency, and facing-angle agreement, with $r_{\mathrm{ori}}^i=(1+\cos\Delta\theta_i)/2$. 
\vspace{-5pt}
\section{Experiments}
\vspace{-5pt}
\definecolor{OVSRowBlue}{HTML}{F1F6FD}
\definecolor{OVSRowGreen}{HTML}{EFF8EE}
\subsection{Experimental Settings and Evaluation Benchmarks}
\vspace{-5pt}
\noindent\textbf{Implementation Details.}
OmniView-Space is built on Qwen3-VL-4B~\citep{qwen3-vl}. For the supervised fine-tuning (SFT) stage, we utilize 26K training data (Refer to the Appendix ~\ref{app:sft_data} for details). For the reinforcement learning (RL) phase in both tool-guided reasoning and cognitive-map distillation, the training data is drawn from an 11K subset filtered from SpatialLadder-26K~\citep{li2025spatialladder} and an additional 10K samples from MindCube~\citep{mindcube}. The tool-guided reasoning stage is trained directly with GRPO, without SFT warm-up. Cognitive-map distillation uses both SFT and RL: we first perform SFT for 3 epochs with a learning rate of $1\times10^{-5}$, and then apply GRPO for 2 epochs with a smaller learning rate of $1\times10^{-6}$ for stable policy updates. We sample $K=8$ trajectories to compute group-relative advantages. All training is conducted on 8 GPUs, each equipped with 90GB of memory.

\noindent\textbf{Evaluation Settings.}
We evaluate OmniView-Space on six spatial reasoning benchmarks covering multi-view aggregation, viewpoint-dependent reasoning, perspective transformation, and embodied spatial cognition. For multi-image reasoning, we report main results on MindCube-Tiny~\citep{mindcube}, MMSI-Bench~\citep{mmsi}, SPAR-Bench~\citep{spar}, and SPBench~\cite{li2025spatialladder}. For single-image spatial reasoning, we evaluate on the Allocentric subset of 3DSRBench~\citep{3dsrbench} and the Perspective subset of OmniSpatial~\citep{jia2025omnispatial}. The results are reported as accuracy following~\citep{vsi}, and proprietary models are shown in gray for reference.
\begin{table}[t!]
\centering
\scriptsize
\setlength{\tabcolsep}{3.2pt}
\renewcommand{\arraystretch}{1.05}
\caption{\textbf{Main results on multi-view spatial reasoning benchmarks.} We report accuracy on MindCube-Tiny, MMSI-Bench, and SPAR-Bench. The \textbf{bold} and \underline{underlined} indicate the best and second-best open-source results; proprietary models are shown in \textcolor{gray}{\emph{gray italic}} for reference.}
\label{tab:main_results}
\resizebox{\linewidth}{!}{%
\begin{tabular}{@{}lcccccccccccccc@{}}
\toprule
\multirow{2}{*}{\textbf{Model}} & \multicolumn{4}{c}{\textbf{MindCube-Tiny}} & \multicolumn{5}{c}{\textbf{MMSI-Bench}} & \multicolumn{4}{c}{\textbf{SPAR-Bench}} & \multirow{2}{*}{\textbf{Overall}} \\
\cmidrule(lr){2-5}\cmidrule(lr){6-10}\cmidrule(lr){11-14}
 & Aro. & Rot. & Amo. & All & Pos. & Att. & Mot. & MSR & All & Low & Mid. & High & All & \\
\midrule
\rowcolor{gray!10}\multicolumn{15}{@{}l}{\textbf{\emph{Proprietary Models}}} \\
GPT-4.1~\citep{hurst2024gpt} & \textcolor{gray}{\emph{59.6}} & \textcolor{gray}{\emph{49.5}} & \textcolor{gray}{\emph{47.2}} & \textcolor{gray}{\emph{50.6}} & \textcolor{gray}{\emph{29.3}} & \textcolor{gray}{\emph{33.8}} & \textcolor{gray}{\emph{38.7}} & \textcolor{gray}{\emph{31.3}} & \textcolor{gray}{\emph{31.7}} & \textcolor{gray}{\emph{38.2}} & \textcolor{gray}{\emph{37.0}} & \textcolor{gray}{\emph{43.1}} & \textcolor{gray}{\emph{39.9}} & \textcolor{gray}{\emph{40.7}} \\
GPT-5.2~\citep{singh2025openai} & \textcolor{gray}{\emph{69.2}} & \textcolor{gray}{\emph{95.5}} & \textcolor{gray}{\emph{41.5}} & \textcolor{gray}{\emph{58.4}} & \textcolor{gray}{\emph{42.9}} & \textcolor{gray}{\emph{47.6}} & \textcolor{gray}{\emph{36.0}} & \textcolor{gray}{\emph{40.4}} & \textcolor{gray}{\emph{42.0}} & \textcolor{gray}{\emph{43.7}} & \textcolor{gray}{\emph{41.5}} & \textcolor{gray}{\emph{66.4}} & \textcolor{gray}{\emph{52.6}} & \textcolor{gray}{\emph{51.0}} \\
Gemini-3-Pro~\citep{Gemini-3-Pro} & \textcolor{gray}{\emph{51.6}} & \textcolor{gray}{\emph{54.5}} & \textcolor{gray}{\emph{46.0}} & \textcolor{gray}{\emph{49.0}} & \textcolor{gray}{\emph{42.9}} & \textcolor{gray}{\emph{50.0}} & \textcolor{gray}{\emph{38.0}} & \textcolor{gray}{\emph{40.9}} & \textcolor{gray}{\emph{42.7}} & \textcolor{gray}{\emph{40.6}} & \textcolor{gray}{\emph{21.7}} & \textcolor{gray}{\emph{58.0}} & \textcolor{gray}{\emph{43.1}} & \textcolor{gray}{\emph{44.9}} \\
\midrule
\rowcolor{gray!10}\multicolumn{15}{@{}l}{\textbf{\emph{Open-source General Models}}} \\
InternVL3-8B~\citep{zhu2025internvl3} & 53.6 & 36.5 & 38.1 & 41.5 & 30.8 & 23.8 & 24.6 & 14.6 & 25.7 & 28.1 & 39.3 & 45.0 & 37.3 & 34.8 \\
Qwen2.5-VL-7B-Instruct~\citep{qwen2vl} & 21.4 & 38.8 & 29.5 & 29.3 & 28.4 & 21.5 & 29.3 & 25.8 & 27.1 & 27.6 & 27.6 & 41.8 & 33.8 & 30.1 \\
Qwen3-VL-4B-Instruct~\citep{qwen3-vl} & 41.6 & 34.1 & 20.0 & 35.9 & 34.2 & 32.3 & 26.0 & 23.7 & 27.7 & 30.8 & 33.1 & 40.5 & 35.6 & 33.1 \\
Qwen3-VL-8B-Instruct~\citep{qwen3-vl} & 53.6 & 36.0 & 32.8 & 38.4 & 31.0 & 26.9 & 31.3 & 28.8 & 30.1 & 35.6 & 32.9 & 39.6 & 36.5 & 35.0 \\
\midrule
\rowcolor{gray!10}\multicolumn{15}{@{}l}{\textbf{\emph{Spatial Intelligence Models}}} \\
MindCube-3B~\citep{mindcube} & 67.6 & 34.0 & 51.0 & 51.7 & 23.9 & 24.6 & 18.7 & 25.3 & 23.9 & 16.8 & 22.6 & 23.9 & 20.8 & 32.1 \\
SpatialLadder-3B~\citep{li2025spatialladder} & 50.8 & 35.0 & 43.2 & 43.4 & 30.3 & 23.3 & 16.0 & 21.2 & 25.4 & 24.9 & 29.4 & 42.4 & 32.9 & 33.9 \\
Spatial-MLLM-4B~\citep{wu2025spatial} & 36.0 & 39.0 & 30.5 & 33.4 & 28.5 & 25.4 & 18.0 & 26.3 & 26.1 & 22.3 & 32.4 & 35.2 & 30.4 & 30.0 \\
SpaceR-7B~\citep{ouyang2025spacer} & 49.2 & 35.0 & 34.2 & 37.9 & 29.1 & 29.4 & 21.9 & 22.5 & 26.9 & 27.0 & 23.9 & 44.6 & 34.2 & 33.0 \\
ViLaSR-7B~\citep{ViLaSR} & 44.4 & 35.5 & 31.0 & 35.0 & 35.9 & 26.0 & 21.0 & 23.2 & 29.8 & 35.3 & 29.0 & 42.5 & 37.4 & 34.1 \\
Cambrian-S-7B~\citep{yang2025cambrian} & 50.8 & 37.0 & 35.9 & 39.7 & - & - & - & - & 27.1 & 29.7 & 31.3 & 49.6 & 37.9 & 34.9 \\
\midrule
\rowcolor{gray!10}\multicolumn{15}{@{}l}{\textbf{\emph{Tool-Integrated}}} \\
\rowcolor{OVSRowBlue}{OmniView-Space-Zero} & {53.6} & 32.5 & {52.8} & {49.1} & 36.6 & {22.3} & 27.3 & 28.4 & \underline{31.7} & {33.5} & {35.8} & {46.3} & {39.6} & 40.1 \\
\rowcolor{OVSRowBlue}{OmniView-Space-R1\textsuperscript{$\dagger$}} & 68.0 & 53.0 & 79.2 & \textbf{71.5} & 41.2 & 35.4 & 28.7 & 25.8 & \textbf{35.5} & 44.1 & 38.1 & 47.8 & \textbf{44.8} & \textbf{50.6} \\
\midrule
\rowcolor{gray!10}\multicolumn{15}{@{}l}{\textbf{\emph{Tool-Distillation}}} \\
\rowcolor{OVSRowGreen}{OmniView-Space-SFT} & {55.8} & {47.0} & {49.8} & {50.9} & {32.6} & {31.5} & 22.7 & 25.8 & {29.6} & {26.6} & {32.5} & {38.9} & {33.1} & 37.9 \\
\rowcolor{OVSRowGreen}{OmniView-Space-R1\textsuperscript{$\ddagger$}} & 71.2 & 48.5 & 69.2 & \underline{65.7} & 36.2 & 31.5 & 24.7 & 24.7 & 31.6 & 43.2 & 34.3 & 44.2 & \underline{42.3} & \underline{46.5} \\
\bottomrule
\end{tabular}%
}
\vspace{-15pt}
\end{table}
\begin{table}[t!]
\centering
\scriptsize
\setlength{\tabcolsep}{2pt}
\renewcommand{\arraystretch}{1.1}
\begin{minipage}[t!]{0.48\linewidth}
\centering
\caption{{Results on the Allocentric subset of 3DSRBench.} \textbf{Bold} and \underline{underlined} numbers indicate the best and second-best performance.}
\label{tab:3dsrbench_allo}
\resizebox{\linewidth}{!}{%
\begin{tabular}{@{}lcccc@{}}
\toprule
\textbf{Model} & Front. & Left. & Twd. & \textbf{Avg} \\
\midrule
RoboBrain2.0-32B~\citep{team2025robobrain} & 60.17 & 38.54 & \underline{27.92} & \underline{42.21} \\
Spatial-SSRL-7B~\citep{liu2025spatial} & 57.41 & 33.95 & 21.72 & 37.69 \\
Cosmos-Reason1-7B~\citep{azzolini2025cosmos} & 60.61 & \underline{40.40} & 21.65 & 40.89 \\
SenseNova-SI-1.1-7B~\citep{cai2025scaling} & 56.10 & 34.81 & 22.96 & 37.96 \\
SpatialThinker-7B~\citep{spatialthinker} & 58.14 & 34.81 & 21.65 & 38.20 \\
REVPT-7B~\citep{zhou2025reinforced} & \underline{61.05} & 35.24 & 21.94 & 39.41 \\
\midrule
\rowcolor{OVSRowBlue}{OmniView-Space-R1} & \textbf{67.15} & \textbf{48.71} & \textbf{30.03} & \textbf{48.63} \\
\bottomrule
\end{tabular}%
}
\end{minipage}
\hfill
\scriptsize
\setlength{\tabcolsep}{3pt}
\renewcommand{\arraystretch}{1.1}
\begin{minipage}[t!]{0.48\linewidth}
\centering
\caption{{Results on the Perspective subset of OmniSpatial.} \textbf{Bold} and \underline{underlined} numbers indicate the best and second-best performance.}
\label{tab:omnispatial_perspective}
\resizebox{\linewidth}{!}{%
\begin{tabular}{@{}lcccc@{}}
\toprule
\textbf{Model} & Ego. & Allo. & Hypo. & \textbf{Avg} \\
\midrule
SpaceMantis-13B~\citep{7} & 49.22 & 38.25 & 39.28 & 42.25 \\
SpaceR-7B~\citep{ouyang2025spacer} & {-} & - & {-} & 41.35 \\
SpaceThinker-7B~\citep{spatialthinker} & 58.04 & 35.11 & 31.08 & 41.41 \\
SpatialBot-3B~\citep{6} & 47.06 & 36.17 & 37.35 & 40.19 \\
V2LO-7B~\citep{huang2025video2layout} & {-} & - & {-} & 46.70 \\
RoboPoint-13B~\citep{yuan2024robopoint} & 49.02 & 37.66 & 33.49 & 40.06 \\
\midrule
\rowcolor{OVSRowBlue}{OmniView-Space-R1} & \textbf{63.73} & \textbf{42.02} & \textbf{57.83} & \textbf{54.53} \\
\bottomrule
\end{tabular}%
}
\end{minipage}
\vspace{-15pt}
\end{table}
\vspace{-15pt}
\subsection{Main Results}
\vspace{-5pt}
\noindent\textbf{Obs 1. Egocentric cognitive maps improve spatial reasoning across input settings.}
Table~\ref{tab:main_results} shows that OmniView-Space outperforms all open-source baselines on three multi-image benchmarks, scoring 71.5 on MindCube-Tiny, 35.5 on MMSI-Bench, and 44.8 on SPAR-Bench. The significant gains on MindCube-Tiny, which requires re-anchoring multi-view observations to query-specific viewpoints, suggest that egocentric cognitive maps provide a more effective reasoning interface than raw images or text alone. MMSI-Bench shows improved robustness across position, attribute, motion, and multi-step questions, while SPAR-Bench confirms the utility of explicit spatial evidence across difficulty levels. The tool-distilled model reaches 46.5 overall, compared with 50.6 for the tool-integrated pipeline, suggesting that distillation retains part of the tool-guided gain. Beyond multi-image reasoning, Tables~\ref{tab:3dsrbench_allo} and~\ref{tab:omnispatial_perspective} evaluate single-image spatial reasoning, where models infer viewpoint-sensitive relations from one observation. OmniView-Space achieves the highest open-source average on the 3DSRBench allocentric tasks and leads on the Allo., Hypo., and Avg. metrics of OmniSpatial, while ranking second on Ego. These results indicate that egocentric map supervision improves internal spatial abstraction rather than only serving as a multi-view fusion mechanism.

\noindent\textbf{Obs 2. Egocentric maps excel in re-anchoring and spatial relations.}
The fine-grained performance breakdown highlights the strengths of OmniView-Space. On MindCube-Tiny, the model excels in the Around, Rotation, and Among subtasks, which directly benefit from egocentric maps due to their reliance on complex relative positioning. On MMSI-Bench, it leads in Position and Attribute categories and remains competitive in Motion and Multi-step reasoning, indicating that explicit spatial grounding supports object-centric decisions. The SPAR-Bench results show similar utility across difficulty levels. Single-image results follow this trend: the model performs best on 3DSRBench (Front, Left, Towards) and OmniSpatial (Allo., Hypo.), which all demand scene transformation into query-specified frames. These consistent gains confirm that egocentric maps are highly effective when tasks require dynamic re-anchoring, spatial comparison, or structuring.
\begin{figure}[t!]
\centering
\includegraphics[width=\linewidth]{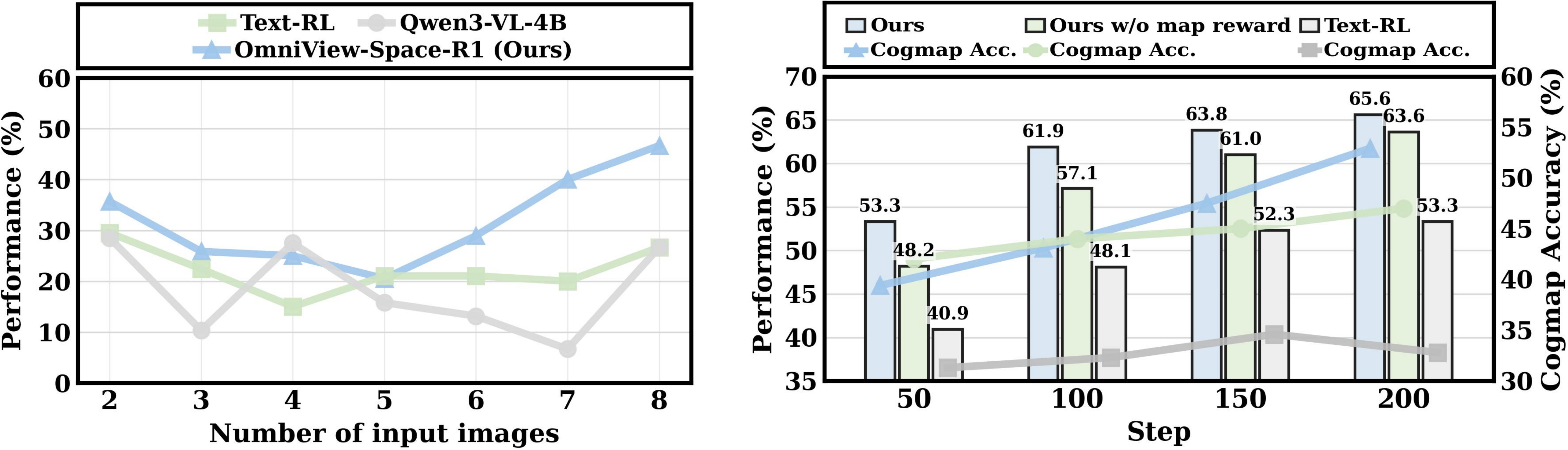}
\vspace{-15pt}
\caption{(a) Accuracy on MMSI-Bench by the number of input images. Tool-integrated RL remains more stable as views increase. (b) Training dynamics of answer accuracy and BEV quality. Image-cogmap supervision improves both metrics more consistently than text-cogmap supervision.}
\vspace{-12pt}
\label{fig:tool_stability_distribution}
\end{figure}
\begin{figure}[t!]
\centering
\includegraphics[width=0.99\linewidth]{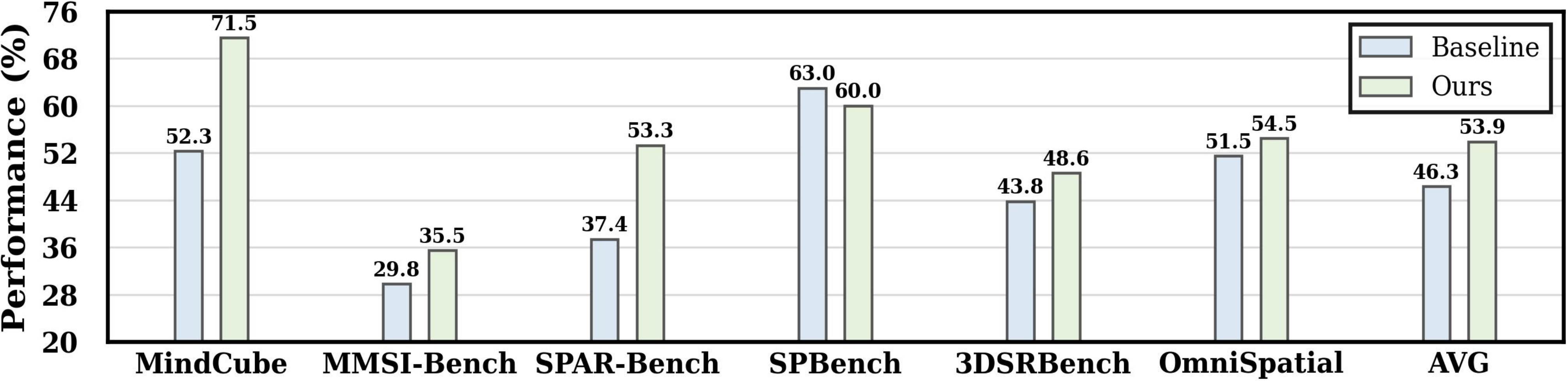}
\vspace{-5pt}
\caption{\textbf{Ablation across six benchmarks.} OmniView-Space broadly improves over the baseline.}
\label{fig:ablation_comparison}
\vspace{-20pt}
\end{figure}

\vspace{-9pt}
\subsection{Ablation Studies}
\vspace{-5pt}
\noindent\textbf{Egocentric Cognitive Map vs. Text-Only Reasoning Baseline.}
Fig.~\ref{fig:ablation_comparison} compares reasoning using multimodal egocentric cognitive maps against text-only approaches. OmniView-Space averages 53.9\%, outperforming the text-only baseline of 46.3\%. This 7.6-point improvement demonstrates that mapping observations into egocentric cognitive maps yields more reliable evidence than relying solely on the textual chain-of-thought. MindCube and SPAR-Bench show the largest gains (19.2 and 15.9 points, respectively), confirming the method's value in tasks requiring viewpoint re-anchoring and structured relational reasoning. The consistent improvements on MMSI-Bench, 3DSRBench, and OmniSpatial suggest this mapping strategy generalizes well to various multi-view settings. The only exception is single-image SPBench, where the text baseline slightly outperforms OmniView-Space (63.0\% versus 60.0\%). This implies that existing language priors can adequately handle single-image tasks with limited viewpoint dependency, reducing the need for explicit spatial mapping.

\begin{wraptable}{r}{0.47\linewidth}
\vspace{-8pt}
\centering
\scriptsize
\setlength{\tabcolsep}{1.1pt}
\renewcommand{\arraystretch}{1.03}
\caption{\textbf{Ablation study.} We isolate the effects of MPSM evidence, tool-guided reasoning, evidence modality, and map-alignment reward.}
\label{tab:mindcube_alignment_ablation}
\resizebox{0.97\linewidth}{!}{%
\begin{tabular}{@{}l@{\hspace{3pt}}c@{\hspace{3pt}}c@{\hspace{3pt}}c@{}}
\toprule
\textbf{Model} & \textbf{MindCube} & \textbf{MMSI} & \textbf{AVG.} \\
\midrule
Qwen3-VL-4B & 35.9 & 27.7 & 31.8 \\
+ MPSM-Map only & 48.4 & 30.3 & 39.4 \\
+ Multimoda MPSM & 49.1 & \textbf{31.7} & 40.4 \\
\midrule
\rowcolor{OVSRowBlue} {OmniView-Space-R1} & \textbf{65.7} & {31.6} & \textbf{48.7} \\
- Map Alignment & 63.6 & 29.1 & 46.4 \\
\bottomrule
\end{tabular}%
}
\vspace{-10pt}
\end{wraptable}
\noindent\textbf{Effectiveness of Multimodal Maps and Map Alignment Reward.}
As shown in Table~\ref{tab:mindcube_alignment_ablation}, first, adding MPSM evidence as prompts improves over Qwen3-VL-4B, showing that query-aligned spatial evidence is useful even without RL training. Second, map-alignment reward improves the distilled model from 46.4 to 48.7 on average, with gains on both MindCube and MMSI-Bench. This supports the role of ego-frame map supervision in producing more useful self-generated cognitive maps.

\noindent\textbf{Robustness to Number of Input Views.}
Fig.~\ref{fig:tool_stability_distribution}(a) reports the MMSI-Bench accuracy grouped by the number of input images. The two baselines, Text-RL and Qwen3-VL-4B, are competitive only for two-view inputs. Their accuracy fluctuates sharply and drops below 10\% when given more than four images, revealing an inability to aggregate evidence across multiple viewpoints. Our tool-integrated RL method is substantially more stable across different image counts. It achieves 35.7\% with two views, maintains 20\% to 30\% for three to six views, and increases to 40.0\% and 46.7\% for seven and eight views. Explicit cognitive-map tool calls appear to convert additional viewpoints into cumulative spatial evidence rather than distracting context, which explains the widening performance gap over the baselines as the input scale grows.
\begin{figure}[t!]
\centering
\includegraphics[width=\linewidth]{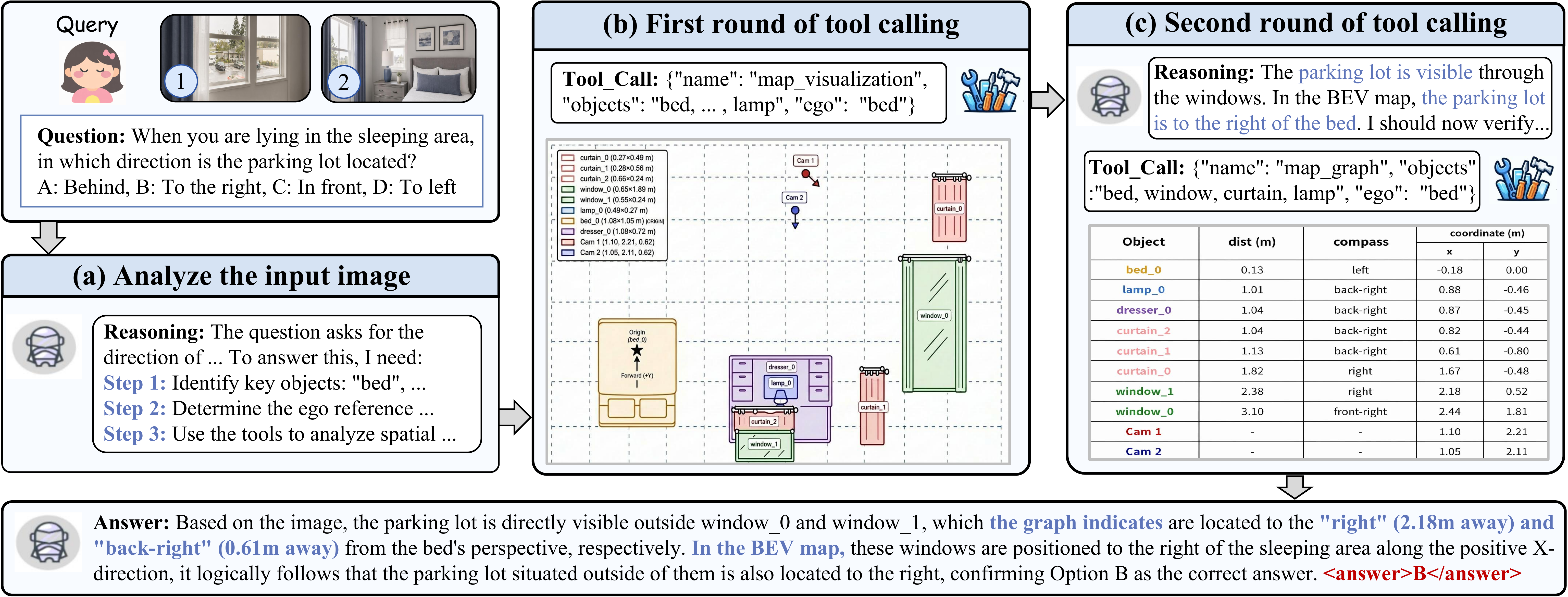}
\caption{\textbf{Case study.} The model first identifies the query anchor and requests a bed-centered BEV map, which makes the window and parking-lot layout directly inspectable. It then calls the graph tool to verify ego-relative relations and grounds the final answer in the query-specified frame.}
\label{fig:reasoning_case}
\vspace{-15pt}
\end{figure}

\noindent\textbf{Cogmap Accuracy and Training Dynamics.}
Fig.~\ref{fig:tool_stability_distribution}(b) compares performance and Cogmap Acc. across training steps. Ours shows the greatest and most consistent improvement, with task performance increasing from 53.3\% to 65.6\% and Cogmap Acc. from 39.4\% to 52.9\%. Ours w/o map reward also improves in performance, from 48.2\% to 63.6\%, but its Cogmap Acc. grows more slowly and is clearly surpassed by Ours in later training. Text-RL remains weaker on both axes: its performance rises only from 40.9\% to 53.3\%, while Cogmap Acc. stays around 31--35\%. These trends indicate that the benefit does not come merely from introducing an intermediate representation. Rendering the egocentric cognitive map as an image gives the model a more faithful spatial state, and the map-alignment reward further links higher-quality self-generated maps with stronger reasoning.

\begin{wrapfigure}{r}{0.46\linewidth}
\vspace{-15pt}
\centering
\includegraphics[width=\linewidth]{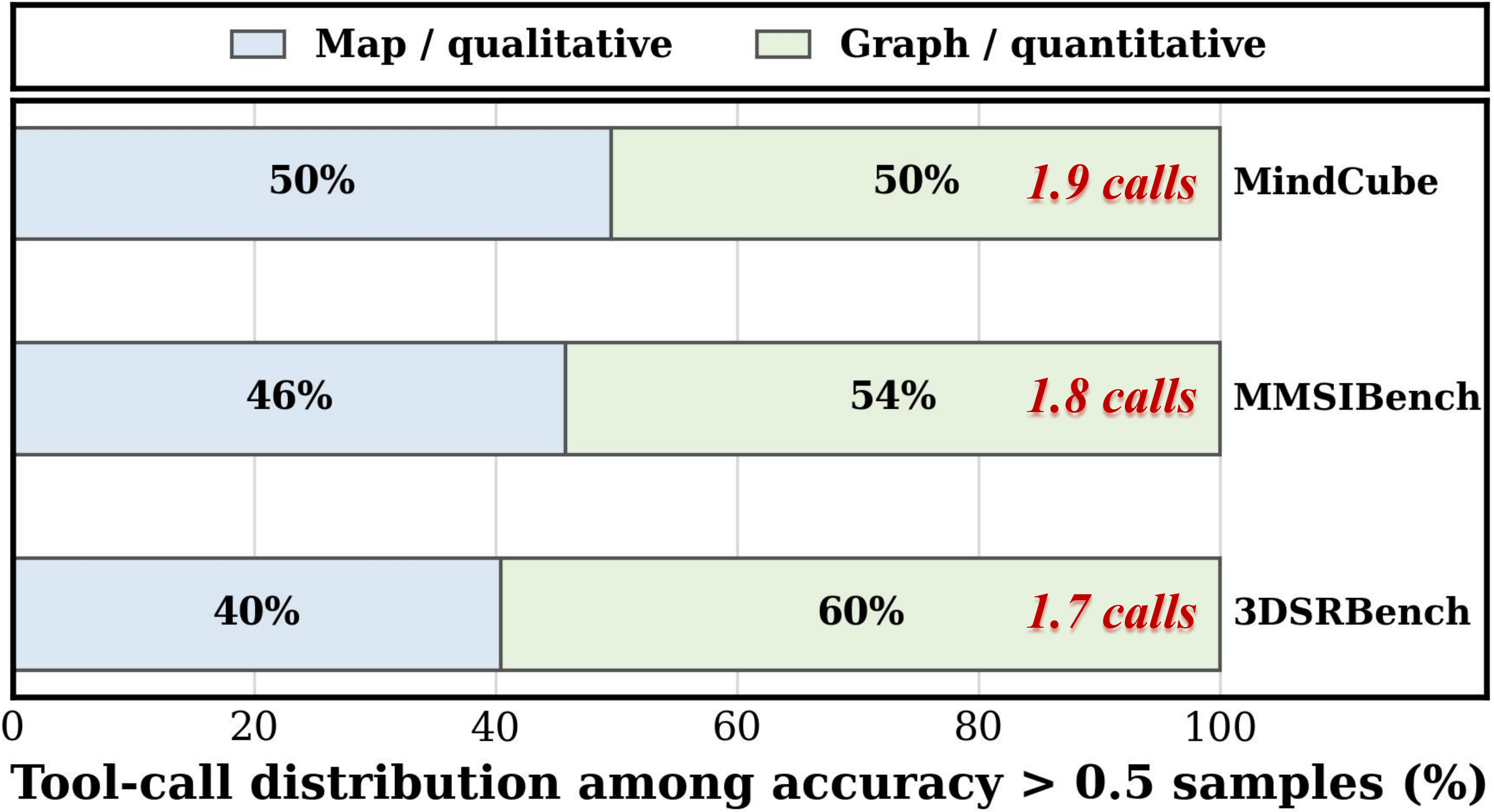}
\caption{{Tool-call distribution across benchmarks.} The Visual Map is preferred for qualitative relative-position questions, whereas the textual graph excels in quantitative queries requiring precise numerical evidence.}
\label{fig:tool_distribution}
\vspace{-10pt}
\end{wrapfigure}
\noindent\textbf{Tool-Call Distribution Analysis.}
Fig.~\ref{fig:tool_distribution} analyzes how our model uses the cognitive map and graph tools on correctly answered samples. MindCube and MMSI-Bench show a balanced use of the two modalities, with image/graph ratios of 50/50 and 46/54, respectively. This matches their emphasis on qualitative viewpoint reasoning, where a visual map provides an intuitive layout for relative-position judgments. By contrast, 3DSRBench places more weight on graph evidence, with graph calls accounting for 60\% of tool usage. This reflects the need for more explicit numerical or structural evidence in detailed 3D relation reasoning. Ultimately, this distribution suggests that different evidence forms are used for different demands.

\noindent\textbf{Qualitative Analysis.}
Fig.~\ref{fig:reasoning_case} illustrates the OmniView-Space reasoning workflow. Based on the input views and spatial question, the model identifies relevant objects and the required reference frame to construct an egocentric cognitive map. This map explicitly represents the target frame, object layout, and relative directions within a visual BEV state. By using this representation as intermediate evidence, the model can accurately evaluate spatial relations from the correct viewpoint. This transparent process enhances interpretability, clarifying the rationale behind specific spatial judgments while mitigating the risk of relying on inconsistent reference frames.
\vspace{-10pt}
\section{Conclusion}
\vspace{-10pt}
We present OmniView-Space, a framework for multimodal egocentric reasoning that shifts spatial analysis from a default system perspective to a query-specified ego frame. To achieve this, we first propose the MPSM module constructs query-aligned visual BEV cognitive maps and textual spatial graphs. We further introduce a tool-guided policy that selects the required ego anchor and grounds answers based on MPSM. Finally, we distill this pipeline using MPSM trajectories and map rewards, enabling the model to reason through internally generated maps. Results confirm that query-aligned egocentric evidence effectively bridges metric geometry and language-based spatial reasoning.

\bibliographystyle{unsrtnat}
\bibliography{references}

\newpage
\appendix
\section{Appendix Overview}
This appendix provides supplementary implementation details to support the main paper. Section~\ref{app:sft_data} elicits the detailed construction, filtering, and repair pipeline for the MPSM-guided SFT data. Section~\ref{app:limitations} discusses the limitations of our approach and highlights future research directions.

\section{MPSM-guided SFT Data Construction}
\label{app:sft_data}
The goal of the SFT data construction pipeline is to teach the model not only to answer spatial questions, but also to externalize an intermediate ego-frame cognitive map before answering. Each full cogmap trajectory is organized as:
\begin{equation}
\label{eq:app_sft_format}
\begin{aligned}
&\texttt{<think>}~R_1~\texttt{</think>}
\rightarrow
\texttt{<cogmap>}~J~\texttt{</cogmap>} \\
&\rightarrow
\texttt{<think>}~R_2~\texttt{</think>}
\rightarrow
\texttt{<answer>}~\hat a~\texttt{</answer>}.
\end{aligned}
\end{equation}
where $J$ is a JSON-format ego-frame cogmap, $R_1$ explains how the cogmap is constructed from the images and the query, and $R_2$ performs forward reasoning from the cogmap to the final answer. We use Qwen3-VL-235B~\citep{qwen3-vl} to produce the language rationales and MPSM to produce the intermediate visual/spatial evidence.

\subsection{Initial Cogmap Trajectory Generation}
For each raw sample with question $q$, image set $\mathcal{I}$, and ground-truth answer $a^{\ast}$, the pipeline first extracts task-relevant object hints and determines the query-specified ego reference. The ego reference can be camera-centric, object-centric, or direction-centric, following the same representation used by MPSM. The object hints and ego reference are then passed to the unified MPSM pipeline, which combines dense geometry, segmentation, and orientation estimation to generate a rendered BEV map and an AST-style spatial tree. The spatial tree is serialized into a JSON cogmap $J$ containing the ego reference, object entries with image-space boxes and BEV footprints, and camera entries with BEV positions and facing angles.

After obtaining $J$, we generate two rationales. The first rationale $R_1$ explains the map construction process: which objects are relevant, why the selected ego frame is appropriate, and how the generated cogmap represents the scene. The second rationale $R_2$ is generated in a forward-reasoning mode: the model is given the question, images, and generated cogmap, and it derives a predicted answer $\hat a$ without being shown the ground-truth answer. The final supervised example is assembled in the format of Eq.~\ref{eq:app_sft_format}. This construction is used for MindCube and SpatialLadder-derived samples, with dataset-specific preprocessing for extracting object hints from grounding annotations, spatial questions, or existing answers.

\subsection{Three-round Filtering and Repair}
The automatically generated trajectories can contain noise from segmentation failures, missing objects, imperfect ego-reference selection, or rationales that reach the right answer through shortcuts. We therefore use a three-round filtering and repair procedure for the full cogmap trajectories.

\noindent\textbf{Round 1: forward-correct filtering.}
We first keep samples whose predicted answer $\hat a$ is already correct under the task metric. For multiple-choice questions, this requires exact option-letter agreement. For numerical questions, we use a thresholded relative-error score. These samples are retained directly because their rationales are generated in a forward manner from cogmap evidence to answer.

\noindent\textbf{Round 2: answer-guided repair with critic filtering.}
For samples that fail Round 1, we ask whether the generated cogmap contains sufficient information to derive the ground-truth answer. If the cogmap is insufficient, the sample is deferred to Round 3. If the cogmap is sufficient, we generate a repaired $R_2$ conditioned on the ground-truth answer, but explicitly require the rationale to remain forward-style rather than saying that the answer is correct by assumption. A separate critic then checks for answer leakage, backward reasoning, circular reasoning, shortcut logic, and factual errors. Only critic-approved repaired trajectories are retained.

\noindent\textbf{Round 3: missing-object refinement.}
For samples whose cogmap is insufficient, the most common failure mode is missing task-relevant objects. We identify objects mentioned by the answer options or question but absent from the cogmap, and use Qwen3-VL-235B to self-ground these missing objects with normalized 2D boxes. These supplemental detections are added to the original object hints, MPSM is rerun to produce an updated BEV map and spatial tree, and $R_1$, $R_2$, and $\hat a$ are regenerated. A sample is retained only if the regenerated answer is correct. This step recovers examples where the original MPSM reference was incomplete due to segmentation or object-filtering noise.

\subsection{Quality Checks}
After construction, we apply lightweight structural and accuracy checks before training. The format checker verifies the presence and nesting of the required tags, including \texttt{<think>}, \texttt{<cogmap>}, and \texttt{<answer>}. It also verifies that the cogmap JSON can be parsed and contains valid object and camera fields. For image-form trajectories, we check that input-image placeholders and rendered-BEV observations are consistent with the conversation structure. We also remove trajectories whose final answer fails the task metric, and flag abnormal samples with missing images, malformed boxes, repeated text, empty cogmaps, or excessive response length. These checks are used to reduce formatting noise before SFT.
\begin{table}[h]
\centering
\small
\caption{\textbf{SFT data mixture.} We combine direct-answer samples with MPSM-guided cogmap trajectories from SpatialLadder, MindCube, and ViewFusion.}
\label{tab:app_sft_mix}
\begin{tabular}{@{}p{0.28\linewidth}p{0.16\linewidth}p{0.44\linewidth}@{}}
\toprule
\textbf{Source} & \textbf{Size} & \textbf{Format / Filtering} \\
\midrule
SpatialLadder grounding & 5K & Learn to predict the BEV Map directly from the input image through \texttt{<think>} $R_1$ \texttt{</think>} $\rightarrow$ \texttt{<cogmap>} $J$ \texttt{</cogmap>} trajectories. \\
SpatialLadder spatial & 7K & Multi-turn reasoning trajectories filtered from SpatialLadder~\citep{li2025spatialladder} after three-round filtering.\\
MindCube \$ VST & 14K & Multi-turn reasoning trajectories filtered from mindcube~\citep{mindcube} and VST-500K~\citep{vst} after three-round filtering. \\
\bottomrule
\end{tabular}
\end{table}
\subsection{Final SFT Mixture}
The final SFT mixture combines direct-answer data, grounding-style cogmap data, and full cogmap-reasoning trajectories. Table~\ref{tab:app_sft_mix} summarizes the sources.
Overall, the mixture contains approximately 26K samples. The direct-answer subset maintains answer-level spatial reasoning ability, while the cogmap subsets teach the model to write structured ego-frame maps and to reason from those maps. The three-round filtering procedure is applied to the full cogmap-reasoning trajectories, while the grounding-only SpatialLadder cogmap data is used directly because it supervises map construction rather than final answer derivation.

\section{Limitations}
\label{app:limitations}
One potential limitation is that the tool-augmented pipeline still depends on the quality of depth estimation, segmentation, and object-orientation estimation. Visual truncation, occlusion, or missing task-relevant objects can lead to incomplete egocentric evidence and incorrect spatial judgments. Although the internalized model reduces reliance on external geometry tools at inference time, its self-generated cognitive maps can still drift geometrically, especially for out-of-domain layouts, sparse views, or ambiguous object orientations. These limitations suggest several directions for future work, including more robust open-vocabulary spatial mapping, stronger map-level supervision, and reasoning policies that can detect when their self-generated maps are unreliable. More broadly, we view ego-frame grounding as a step toward MLLMs that not only describe visual scenes, but also re-instantiate them from the viewpoint required by a task.


\end{document}